# LSTM-based Preceding Vehicle Behaviour Prediction during Aggressive Lane Change for ACC Application

Rajmeet Singh[1], Saeed Mozaffari[1], Mahdi Rezaei[2], Shahpour Alirezaee[1]
[1]Mechanical, Automotive, and Material Engineering Department. University of Windsor, Windsor, Canada
{rsbhourji, saeed.mozaffari, s.alirezaee } @uwindsor.ca
[2]Institute for Transport Studies, University of Leeds, UK
m.rezaei@leeds.ac.uk

*Abstract*—The development of Adaptive Cruise Control (ACC) systems aims to enhance the safety and comfort of vehicles by automatically regulating the speed of the vehicle to ensure a safe gap from the preceding vehicle.However, conventional ACC systems are unable to adapt themselves to changing driving conditions and drivers' behavior. To address this limitation, we propose a Long Short-Term Memory (LSTM)based ACC system that can learn from past driving experiences and adapt and predict new situations in realtime.The model is constructed based on the real-world *highD* dataset, acquired from German highways with the assistance of camera-equipped drones. We evaluated the ACC system under aggressive lane changes when the side lane preceding vehicle cut off, forcing the targeted driver to reduce speed. To this end, the proposed system was assessed on a simulated driving environment and compared with a feedforward Artificial Neural Network (ANN) model and Model Predictive Control (MPC) model. The results show that the LSTM-based system is 19.25 % more accurate than the ANN model and 5.9 % more accurate than the MPC model in terms of predicting future values of subject vehicle acceleration. The simulation is done in Matlab/Simulink environment.

## I. INTRODUCTION

Despite the growing number of vehicles on the roads, the matter of ensuring road safety is frequently disregarded.The predominant form of road traffic accident is a rear-end collision, which arises when a vehicle collides with the one ahead of it. This type of collision constitutes 29% of all crashes and is responsible for 7.2% of fatalities [1], with the majority of these incidents attributed to human error.Advanced driver-assistance systems (ADAS) have been created to improve safety and driving comfort by providing alerts, warnings, assistance, or taking control of the vehicle when necessary [2]. ACC is a crucial element of ADAS, as it enables the longitudinal control system to automatically regulate the velocity of the vehicle and sustain a secure gap between the host and the preceding vehicle [3]. Several control approaches have been investigated in the extensive research on ACC, including the PID-based control [4], fuzzy logic control [5], model predictive control (MPC) [6], and neural networks (NN) [7]. The utilization of MPC in ACC systems offers several advantages, including its ability to achieve precise and optimal control, real-time multi-objective optimal control, and even high responsiveness during traffic congestion [8]. Deep learning has become a prominent focus of research in numerous systems and applications in recent years and has been used in various transportation and autonomous driving applications such as ACC systems [9, 10], cooperative adaptive cruise control (CACC) [11], traffic sign recognition [12], and map merging [13]. As per the author's knowledge, there was no use of the real-world data set in the above literature for developing the models.

This paper aims to extend the state-of-the-art by providing a data-driven approach for predicting the behavior of preceding vehicles and incorporating it into the ACC. The main contribution of this paper is to predict the ACC parameter (acceleration m/s$^2$) of the subject vehicle (SV) during aggressive lane change by the side lane preceding vehicle (PV), and adapt the ACC parameters for a safer driving experience. Toward this aim, highly disaggregated naturalistic driving data from the *highD* dataset are utilized [14].The dataset comprises over 45,000 km of naturalistic driving behavior, which was derived from 16.5 hours of video footage captured by camera-equipped drones on German highways. A Matlab-based program was initially employed to identify patterns of aggressive lane changes by the preceding vehicle in relation to the host vehicle in the data. The resulting data was then utilized as input for a real-time long short-term memory (LSTM) deep neural network, which predicted the desired acceleration of the host vehicle to adhere to the ACC scenario.

The paper is organized into several sections. Firstly, the problem statement is introduced, which is then followed by the methodology employed in the current study. A description of the data and its pre-processing is provided subsequently. The LSTM approach is then employed to train and test the data. Finally, the results of the proposed model are compared with the ANN model for predicting future acceleration values during aggressive lane change by the side lane preceding vehicle and conclusions drawn.

## II. PROBLEM STATEMENT

We consider the problem of adapting the ACC parameter (acceleration m/s$^2$) of the subject vehicle in advance to avoid aggressive lane changes by the preceding vehicle driving on a highway, using previously observed data. Aggressive lane change refers to a cut-off situation where a PV driver changes lanes so closely in front of SV that the driver must reduce speed suddenly to avoid a collision. Fig. 1 shows the scenarios when the preceding vehicle aggressively changes the lane from 3 to 2. Formally, we consider a set of observable features $\Psi$ and a set of target outputs $\Theta$ to be predicted.

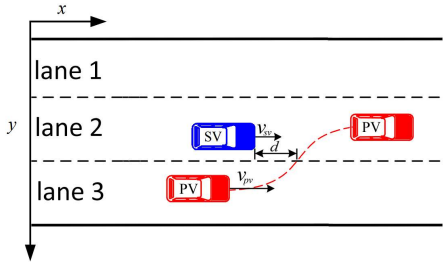

Figure 1. The diagram of ACC parameter decision-making during the aggressive lane change process.

In this article, our approach is to train a predictor for the future acceleration of the SV and incorporating it during ACC implementation. Therefore, we limit the amount of available information to the vehicles immediately in front of the SV which can be captured in the ACC application. LSTM had been used to train and test the processed data from the *highD* dataset [14].

### III. DATA AND FEATURES

*A. Dataset*

The *highD* dataset [14] consists of German highways information with the assistance of camera-equipped drones. Traffic was recorded at six different locations and includes more than 110500 vehicles. Although the dataset was mainly created for the safety validation of highly automated vehicles, it is also suitable for many other tasks such as the analysis of traffic patterns or the parameterization of driver models. The dataset includes vehicles' ID, position, speed, acceleration, etc. shown in Table I.

*B. Data Screening*

In this paper, the *highD* data during the aggressive lane change by the preceding vehicle is counted. Based on the analysis of the factors affecting the ACC, combined with the decision-making behavior of the subject vehicle acceleration in the actual driving situation, a total of 5 data of two vehicles (subject and preceding vehicles) in a driving unit are selected as the ACC decision parameters. Based on this information, we aim to predict the acceleration of SV as the output of the ACC model.

1) $X_{sv}$ = position of the SV
2) $X_{pv}$ = position of the PV
3) $V_{sv}$ = velocity of the SV
4) $V_{pv}$ = velocity of the PV
5) $d$ = distance between SV and point of lane change by PV
6) $ACC_{sv}$ = acceleration of the SV

When the SV encounters a situation where the side lane PV changes lanes and cuts in front of it, the SV driver will decrease speed and apply the brakes. Once the PV has passed, the SV will resume following the adaptive cruise control (ACC) system and maintain a safe distance from the PV.

TABLE I. MAIN CHARACTERISTICS PARAMETERS PROVIDED BY THE HighD DATASET

| Number | Data name | Unit |
|---|---|---|
| 1 | Current frame. | number |
| 2 | Track's id | number |
| 3 | X position | m |
| 4 | Y position | m |
| 5 | Width of vehicle | m |
| 6 | Height of vehicle | m |
| 7 | X_velocity | m/s |
| 8 | Y_velocity | m/s |
| 9 | X_acceleration | $m/s^2$ |
| 10 | Y_acceleration | $m/s^2$ |
| 11 | Preceding X_velocity | m/s |
| 12 | Lane_Id | number |

The decision parameter (acceleration) is extracted by pre-processing and filtering the highD dataset, and some of the sample data are shown in Table II.

### IV. LONG SHORT-TERM MEMORY NETWORK MODEL

Recurrent Neural Networks (RNNs) are a unique type of neural network composed of multiple neural networks linked together in a chain-like structure, allowing them to model temporal dependencies in a sequence [15]. However, in practice, RNNs can struggle to model long dependencies effectively [16]. To overcome this challenge, a specialized form of RNN called Long Short-Term Memory (LSTM) was developed [17]. LSTMs retain the chain-like structure of RNNs but with modified individual units that enable them to learn long-term dependencies more effectively. LSTM networks employ input, forget, and output gate layers in addition to a memory cell, allowing them to regulate the flow of information. These gate layers are responsible for discarding non-essential information and retaining only the essential information required for a given task. LSTMs have been utilized in predicting highway trajectories, determining driver intentions at intersections, and lane change maneuvers [18]. To develop a decision model for a vehicle's ACC system, this paper utilizes LSTM networks. The LSTM model is constructed using MATLAB software.

*A. Training and test data*

The dataset covers both four-lane (two per direction) and six-lane (three per direction) highways with central dividing medians and hard shoulders on the outer edge. The recordings were made on highways. The dataset contains data from 110,000 vehicles (81% cars and 19% trucks), covering a total distance of 45,000 km, with 5,600 lane changes observed. We considered 40 frames before and after aggressive lane change. The current study uses only the longitudinal velocity and longitudinal acceleration time series from the dataset. Only the trajectory data from recording numbers 9 to 60 are used to train and prediction models. This selected data includes 2,449 vehicles (2,034 cars and 415 trucks) recorded, of which 259 vehicles executed aggressive lane changes. Including both cars and trucks in the dataset allows for the capture of driver behavior in mixed traffic scenarios. The data is divided into training and validation set using an 80/20 ratio, shown in Table III. The test data set comprises 15,432 trajectories.

TABLE II. ACC MODEL DECISION FACTOR DATA

| $X_{sv}(m)$ | $X_{pv}(m)$ | $V_{sv}(m/s)$ | $V_{pv}(m/s)$ | $d(m)$ | $ACC_{sv}(m/s^2)$ |
|---|---|---|---|---|---|
| 154 | 189.16 | 24.93 | 30.78 | 18.08 | 0.03 |
| 78.71 | 115 | 24.98 | 32.89 | 23.79 | 0.035 |
| 122.58 | 162.9 | 23.41 | 26.37 | 20.51 | 0.05 |
| 209.92 | 252.89 | 30.61 | 34.24 | 35.17 | 0.06 |
| 213.58 | 256.64 | 30.62 | 34.24 | 35.78 | 0.07 |
| 222.12 | 266.22 | 30.64 | 34.23 | 36.82 | 0.08 |
| 224.56 | 268.96 | 30.65 | 34.24 | 37.12 | 0.09 |

TABLE III. TRAINING SAMPLES AND TEST SAMPLES

| Dataset | Aggressive lane change followed by the subject vehicle |
|---|---|
| Training data | 12345 |
| Testing data | 3087 |
| Total | 15432 |

*B. Determination of LSTM predictor*

To determine the optimal configuration of the LSTM predictor, its performance on the validation dataset and training time is assessed. The flow chart of the proposed network layers is shown in Fig.2. It consists of eight layers each has 200 neurons. A sequence input layer inputs sequence data to a network. A fully connected layer multiplies the input by a weight matrix and then adds a bias vector.The ReLU layer performs a threshold operation on each element of the input, where any value less than zero is set to zero.An LSTM layer learns long-term dependencies between time steps in time series and sequence data.The layer performs additive interactions, which can help improve gradient flow over long sequences during training and the regression layer computes the half-mean-squared-error loss for regression tasks. The Adam optimizer is utilized to adjust the learning rate [19]. The learning rate is set to 0.0001. The training process is stopped when the validation accuracy does not show improvement over five consecutive iterations/epochs to prevent overfitting.

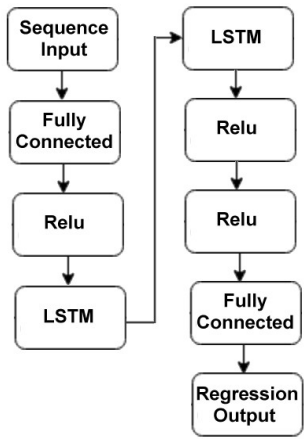

Figure 2. Proposed network layers

V. RESULTS

The performance of the vehicle acceleration prediction is evaluated in this section based on root mean square error (RMSE) as an indicator of the model prediction accuracy which is formulated as the following equation (1) and (2) [18]:

$$\bar{y} = \frac{1}{N}\sum_{t=1}^{N} y_t \qquad (1)$$

$$\text{RMSE} = \sqrt{\frac{\sum_{t=1}^{N}\left(y_t - \hat{y}_t\right)^2}{N}} \qquad (2)$$

where $\bar{y}$ is the mean of the measured acceleration, $y_t$ is the measured acceleration, $\hat{y}_t$ is the predicted acceleration, and $N$ is the number of elements in output. Fig. 3 shows the RMSE values during the LSTM training process. As depicted in Fig. 3 the value of RMSE converges very fast and reaches below 0.025 values in 2000 iterations.

The proposed model is compared to the feedforward Artificial Neural Network (ANN) with five hidden layers each has 200 neurons and Model predictive control (MPC) [6] methods for prediction accuracy. The results are presented in Fig. 4, which showed that the future predicted values of the proposed model are more accurate than those of the ANN and MPC model. The overall prediction accuracy of the proposed model was found to be 98.5%, which is significantly higher than the accuracy of the ANN model (79.25 %), and MPC model (92.6 %). Hence proposed model performs better for the stated scenario.

VI. ACKNOWLEDGMENT

We acknowledge the financial support from the Natural Sciences and Engineering Research Council of Canada (NSERC) Catalyst Grant. Also, the assistance provided by Mrs. Manveen Kaur (University of Windsor) in utilizing a Matlab program to refine the raw data is acknowledged by the authors.

VII. CONCLUSIONS

In this study, an LSTM model was proposed to predict the ACC future values of a subject vehicle's acceleration during aggressive lane change caused by a side lane preceding vehicle. The proposed system utilizes a comprehensive dataset acquired from German highways and can learn from past driving experiences to adapt and predict new situations in real-time.

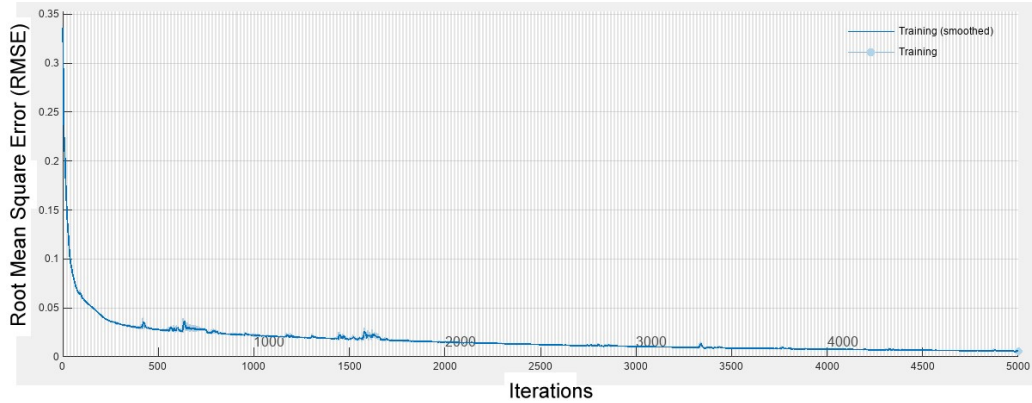

Figure 3. RMSE results.

The proposed LSTM-based system was compared with other methods, and the results show that our system outperforms state-of-the-art methods in predicting the future values of the subject vehicle acceleration. Therefore, it can be concluded that the proposed model is more effective and accurate in predicting the subject vehicle's acceleration during an aggressive lane change by side lane preceding vehicle. For future scope, authors will validate the proposed model for different scenarios such as lane merging, roundabouts, etc.

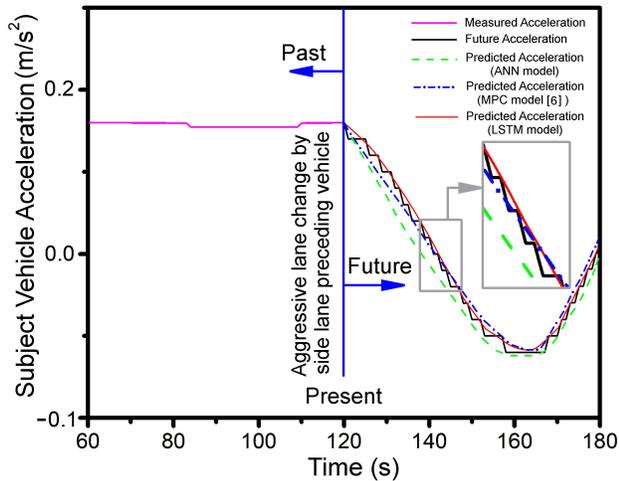

Figure 4. Prediction results of the proposed model.